\documentclass[pdflatex,sn-mathphys-num]{sn-jnl}

\usepackage{graphicx}%
\usepackage{multirow}%
\usepackage{amsmath,amssymb,amsfonts}%
\usepackage{amsthm}%
\usepackage{mathrsfs}%
\usepackage[title]{appendix}%
\usepackage{xcolor}%
\usepackage{textcomp}%
\usepackage{manyfoot}%
\usepackage{booktabs}%
\usepackage{algorithm}%
\usepackage{algorithmicx}%
\usepackage{algpseudocode}%
\usepackage{listings}%
\usepackage{placeins}%

\theoremstyle{thmstyleone}%
\theoremstyle{thmstyletwo}%

\theoremstyle{thmstylethree}%

\begin{document}

\title[LangLingual]{LangLingual: A Personalised, Exercise-oriented English Language Learning Tool Leveraging Large Language Models}

\author*[1]{\fnm{Sammriddh} \sur{Gupta}}\email{sammriddh.gupta@student.unsw.edu.au}

\author[1]{\fnm{Sonit} \sur{Singh}}\email{sonit.singh@unsw.edu.au}

\author[1]{\fnm{Aditya} \sur{Joshi}}\email{aditya.joshi@unsw.edu.au}

\author[2]{\fnm{Mira} \sur{Kim}}\email{mira.kim@unsw.edu.au}

\affil*[1]{\orgdiv{School of Computer Science and Engineering}, \orgname{University of New South Wales}, \orgaddress{\street{High Street}, \city{Kensington}, \postcode{2052}, \state{NSW}, \country{Australia}}}

\affil[2]{\orgdiv{School of Humanities and Languages}, \orgname{University of New South Wales}, \orgaddress{\street{High Street}, \city{Kensington}, \postcode{2052}, \state{NSW}, \country{Australia}}}


\abstract{
While language educators strive to create a rich experience for learners, they may be restricted in the extend of feedback and practice they can provide. We present the design and development of \textit{LangLingual}, a conversational agent built using the LangChain framework and powered by Large Language Models. The system is specifically designed to provide real-time, grammar-focused feedback, generate context-aware language exercises and track learner proficiency over time. The paper discusses the architecture, implementation and evaluation of LangLingual in detail. The results indicate strong usability, positive learning outcomes and encouraging learner engagement. We conclude with a plan for future enhancements. 

}




\maketitle

\section{Introduction}\label{sec1}

English language support is often provided in the form of courses and academic programs to students who use English as a second language. In most educational settings, it is not possible to provide personalised attention to each learner due to time and resource constraints. For a cohort of English language learners, there may be a wide range of proficiency levels and learning needs. Large language models (LLMs) have been identified to potentially aid learning~\cite{Mohamed2024-ax}. We introduce LangLingual, a novel web-based tool that utilises LLMs to create personalised learning plans for English language learners. Proficiency levels of learners are tracked based on rubrics used by a Linguistics Professor who runs a personalised English language learning program. The proficiencies are used to create exercises that the learner answers in a conversational setting while getting feedback at every stage. The feedback is tracked for the learner to see their progress. The versatility of LangLingual is that it can be used by learners of varying proficiency levels.

Although commercial, public LLMs can give answers to English learning questions, they provide a response to learner input and may not track learner progress over multiple sessions. In contrast, a human teacher helps a student to think for themselves and tries to give them hints or resources before providing the answer. We hope for our educational application to also behave like a teacher who encourages and provides hints to students. This refers to Socratic method of learning. The personalised approach can help address specific areas where learners struggle and promote more effective learning, thus enhancing overall English language proficiency.

LangLingual can enhance the learning experience and address individual needs efficiently. It is not limited to a traditional classroom, and can be used by a global audience. This tool can assist teachers to have a greater impact and provide insights into learning trends. This study contributes to the field of education technology by showing how LLMs can be effective learning tools that people all over the world can benefit from.  

Unlike general-purpose LLM applications, LangLingual is a specialised, end-to-end system focused entirely on English language learning. Learners can immediately start improving their English without needing to configure prompts, settings or model instructions manually. LangLingual treats every interaction with finesse, even casual sentences are evaluated from a language improvement perspective, rather than interpreting them for tasks like finding locations or recommendations.
This ensures a consistent educational experience. Additionally, the system has been designed to be largely model-agnostic for future scalability. Although it currently integrates with OpenAI models, the architecture allows for inclusion of other LLM providers in the future to suit learner's preferences. Thus, by combining English language-specific personalisation, and strong focus on educational application, LangLingual differentiates itself from the myriad of generic LLM tools available today.

The rest of the paper is organised as follows. Section~\ref{sec:relwork} presents past work in language learning tools using LLMs. Following that, we present the architecture of LangLingual in Section~\ref{sec:arch}.

\section{Related Work}\label{sec:relwork}

The landscape of language learning technologies has evolved greatly over the years, with advancements in NLP and DL techniques. Studies in developing effective language learning tools highlights both the limitations of current systems in terms of not having personalised learning but also highlighting the need to have innovative methodologies to handle key issues including the need for broader language task coverage, better natural language understanding, scalability, ethical considerations, and the ability to handle mixed-language queries. Studies also highlighted the need to have benchmark datasets and critical evaluation methods to check robustness and generalisability of methods for language learning. Various methods have been proposed, ranging from traditional NLP techniques to DL to hybrid approaches. 

In Argumentative Writing Support To Foster {E}nglish Language Learning (ALEN)~\cite{wambsganss-etal-2022-alen}, authors trained a multi-class classifier at the sentence level, and used SVM for argument component classification and persuasive relation classification. In Bot for {E}nglish Language Acquisition (BELA)~\cite{mahajan-2022-bela}, authors leverage the XLM-RoBERTa model~\cite{papluca/xlm-roberta} for language identification and integrate transformer-based machine translators from salesken.ai. This model is extremely apt in capturing language-specific nuances as well as transliteration and translation tasks. Another innovative method is a custom backend system constructed via various NLP techniques, utilised by the Language Muse~\cite{madnani-etal-2016-language} project. This system processes user-uploaded texts, identifies linguistic features and generates JSON objects that help in creating various language learning exercises. This approach allows the design of an immensely creative and flexible educational tool. Grammar Error Correction (GEC) is a significant NLP task and pivotal in language learning tools, helping to provide real-time feedback on user input. The ChatBack~\cite{liang-etal-2023-chatback} chatbot processes user inputs through a GEC module and switches to a feedback mode when targeted errors are identified. This is an example of an approach where learners receive immediate feedback and can choose to self-correct or receive hints to improve their skills. To develop an 'open-domain chatbot for language practice~\cite{tyen-etal-2022-towards}, authors utilise several strategies to adjust difficulty levels according to the Common European Framework of Reference for Languages (CEFR)~\cite{cefr}. They use models like Facebook's Blender 2.7B~\cite{roller-etal-2021-recipes} and implement various sampling methods such as top-k sampling~\cite{fan-etal-2018-hierarchical} and sub-token penalties~\cite{sennrich-etal-2016-neural}. This ensures the conversational agent can cater to different language proficiency levels. By introducing various datasets, including, EFCAMDAT\cite{efcamdat-modified} and CLC-FCE\cite{clc-fce}, researchers developed scoring models that provide accurate  proficiency assessments. The Innovative use of NLP for Building Educational Applications (BEA) shared task 2023 on generating AI teacher responses~\cite{tack-etal-2023-bea} demonstrated the usage of a variety of state-of-the-art (SOTA) models including Alpaca, Bloom, DialoGPT, DistilGPT-2, Flan-T5, GPT-2~\cite{radford2019language}, GPT-3\cite{brown2020languagemodelsfewshotlearners}, GPT-4, LLaMa, OPT-2.7B and T5-base. The task organisers found that generative pre-trained transformer (GPT) language models are extremely good at educational tasks like dialogue generation, context identification and grammar rules. Fine-tuning and combining these models with sentence-level embeddings via Sentence-BERT~\cite{reimers-gurevych-2019-sentence} allows for nuanced and context-aware interactions. 

To overcome some of the limitations of the LLMs, retrieval-augmented generation (RAG) approach~\cite{Lewis_2020_RAG} has been widely applied. In multilingual Hypothetical Exercise Retriever (myHyER)~\cite{Xu_2024:LLM_augmented_exercise_retrieval}, authors leverage the generative capabilities of LLMS to synthetically generate hypothetical exercises based on the learner's input. This approach reduces the semantic gap as vector similarity approaches poorly captures the relationship between exercise content and the queries that learners use to express what they want to learn. In \cite{Kohnke_2025_exploring}, authors examined the influence of GenAI on L2 learners' language competencies, focusing on tools commonly used by first-year students learning academic English. Authors find that 66.7\% of students regularly use GenAI tools. Students found that GenAI tools helped them in improving grammar, writing, vocabulary, and reading skills. Research also find that students like personalised and creative support provided by the GenAI tools. In \cite{cui-sachan-2023-adaptive}, authors used knowledge tracing models that estimates each student's evolving knowledge states from their learning history to generate adaptive and personalised exercises for language learning. In \cite{Kim_2024_exploring_LLMs}, authors proposed an adaptable LLM-based chatbot to foster growth mindset and aid in language learning. Authors found that mastering language posed several challenges, including motivation and fear of failure. To over the limitations of fixed mindset and to inculcate growth mindset, authors used adaptable LLM, having growth mindset style and user perception. 

Recent work also demonstrates the emerging role of LLM and AI technology in education~\cite{Yadegaridehkordi2025-av}.
\section{Architecture}\label{sec:arch}

The system developed in this works is named \textbf{LangLingual}. The name reflects its focus on language learning and improvement through interactive conversations. LangLingual is designed as a personalised educational assistant that supports learners in practising and refining their English skills through a personalised exercise-oriented approach for conversations. The architecture is shown in Figure~\ref{fig:architecture2}. 

\subsection{Data Sources}

Two primary data sources are used in the functionality of LangLingual.

\begin{itemize}
    \item \textbf{Word Bank:} A comprehensive vocabulary dataset consisting of approximately 50,000 English words labelled with proficiency levels ranging from 1 (beginner) to 14 (advanced). The dataset was pre-processed and transformed into a structured CSV format with two columns: \texttt{[word, level]}. For example, an entry in the dataset may appear as \texttt{[ability, 1]}. These are words used by a linguistics professor with 20 years of experience and who convenes a personalised language learning course at a large, international, research-intensive university.
    \item \textbf{Resources Dataset:} A manually curated collection of learning resources from the internet, compiled to cover various areas of English language improvement. The dataset was stored in CSV format with the following columns: \texttt{[area, resource\_type, title, description, url, difficulty\_level]}. It includes resources targeting topics such as Articles, Adjectives, Adverbs, General Grammar, Tenses, Subject–Verb Agreement, Vocabulary Range, Sentence Structure, Word Order, Phrasal Verbs, Idioms, Conjunctions, Word Choice, Repetition, Punctuation, and Capitalisation.
\end{itemize}

\subsection{Proficiency Assessment}
To estimate the proficiency level of a learner, a hybrid scoring approach is used, which combines rule-based and model-based evaluation.

The first method uses a word bank. The word bank, which is discussed in the data sources section is used here. Each word from the learner input text is lemmatised and checked against the word bank. For matched words, average and median levels are calculated. The second method uses LLM as a judge. Specifically, the same learner input text is passed to an LLM with a carefully crafted prompt, requesting a level prediction between 1-14. The model returns a numeric estimate.

The weights from the two methods are combined to compute proficiency of a learner. In order to do so, we use the following formula:
\[
\text{Level}_{\text{combined}} = w_{\text{wb}} \times \text{Level}_{\text{wb}} +  w_{\text{llm}} \times \text{Level}_{\text{llm}}
\]
where wb = word bank, llm = large language model, w is weight \( w_{\text{wb}} = 0.4 \) and  \( w_{\text{llm}} = 0.6 \)

This hybrid approach ensures more accurate proficiency estimation especially because 50,000 words does not sufficiently reflect all possible words in the English Language.

\subsection{Exercise Generation}
Based on the learner's proficiency level and context, LangLingual detects and generates exercises for users throughout the session. After the LLM generates a message, an exercise generation module checks the assistant reply for exercise-related phrases. If particular keywords are found, that particular message is flagged as an exercise and information is stored in the form of \emph{exercise type} and \emph{exercise prompt}. A dictionary stores details of the active exercise, enabling system to remember if the learner is currently solving an exercise or has moved on to other parts. After the learner has attempted exercises, the LangLingual provides an insightful feedback for the user in terms of what are improvement areas.

\subsection{Improvement Area Identification}
To provide actionable feedback, LangLingual analyses a learner's conversation input to identify key areas where they need to improve. This analysis occurs at repeated interval, after enough learner input has been collected. All learner messages are aggregated into a single chunk of text. If there is sufficient input (a heuristic value of 3 learner interactions), then the text chunk is passed to LLMs for analysis. The focus is on searching for consistent patterns of grammatical or lexical issues, rather than one-off errors. Some example Categories Assessed include Articles, Tenses, and Subject-Verb Agreement. The LLM returns a JSON array with 3 areas of improvement, with each including: (a) The identified improvement area. (e.g. "Articles"); (b) \texttt{confidence}: A float between 0.0 and 1.0 indicating frequency and severity of the issue; (c) \texttt{examples}: A list of learner phrases where issues were found. To ensure quality, only areas with confidence greater than 0.3 are included in the final output.

\section{System Details}
\begin{figure}[htb]
    \centering
    \includegraphics[width=0.9\textwidth]{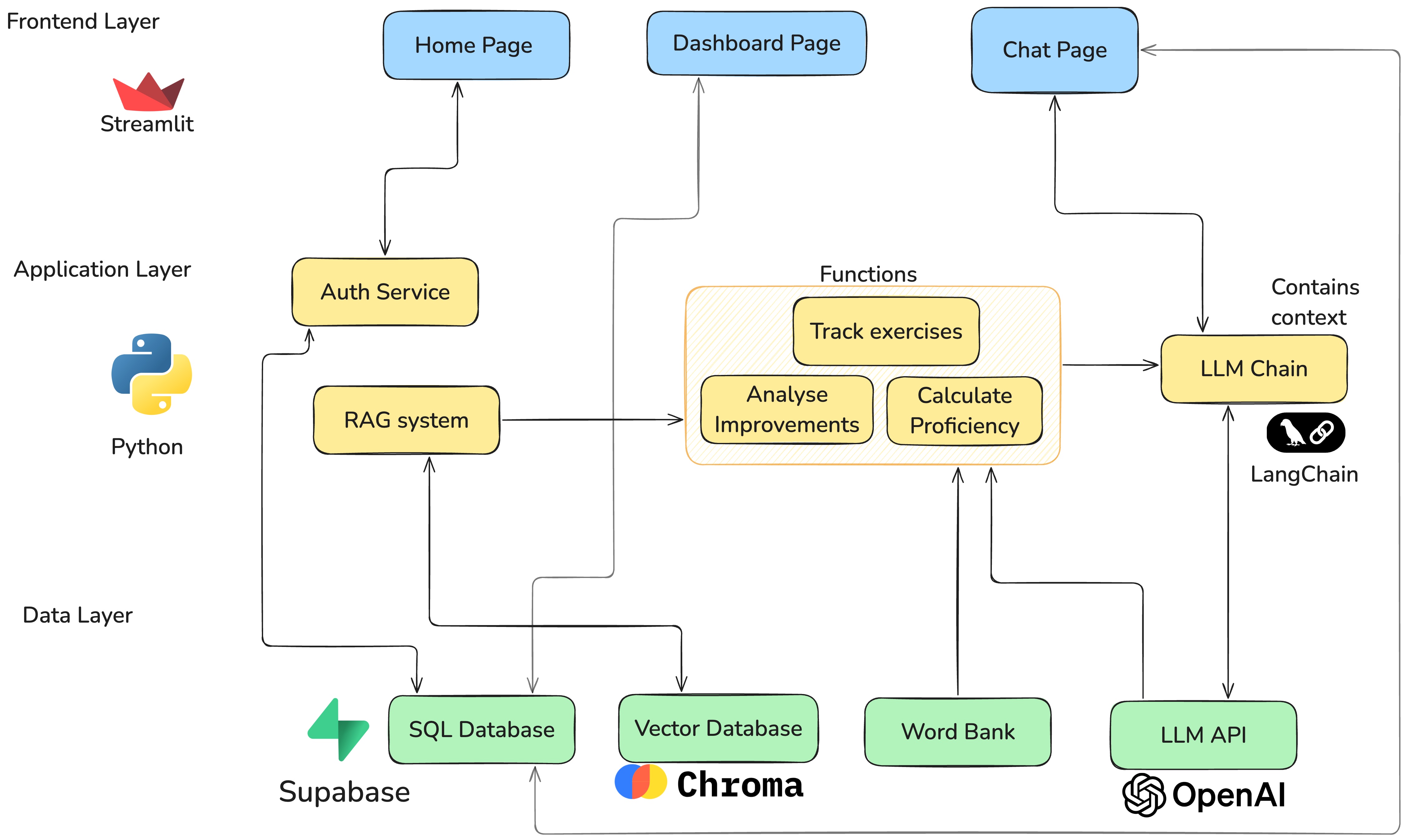}
    \caption{Architecture Diagram of the LangLingual system}
    \label{fig:system}
\end{figure}

The system details of LangLingual are illustrated in Figure\ref{fig:system}. LangLingual was implemented in Python owing to its support for web frameworks, LLMs and database querying. The open-source Python framework \textit{Streamlit}\cite{streamlit}, is used to build the front-end while LangLingual is deployed and hosted on \textit{Streamlit Cloud}\cite{streamlit} which does not require manual servers. Similarly, \textit{Supabase}\cite{supabase} is used for learner-authentication (via email login) and as the backend database. It provides a PostgreSQL\cite{postgres} instance with real-time updates and row-level security policies. \textit{PostgreSQL}\cite{postgres} is used to store learner profiles, session metadata, chat messages and improvement areas. Row-level security ensures learners can only access their own data. These provisions allow to track individualised progress in LangLingual.

In terms of the backend, the open-source framework \textit{LangChain} (v0.3)\cite{langchain} is used to construct the conversational pipeline.  OpenAI's\cite{openai} GPT models\cite{gpt-models} are utilised via LangChain\cite{langchain} for features such as real-time responses, proficiency calculation and generating personalised English grammar exercises. \textit{OpenAI's\cite{openai} Whisper} model API\cite{whisper} handles the speech-to-text transcription for voice inputs. It is used to manage context, memory, prompt-chaining and RAG-based retrieval. Similarly, the open-source vector database \textit{ChromaDB}\cite{chroma} is used to store document embeddings.

\FloatBarrier

\section{Evaluation}\label{sec2}
We evaluate LangLingual using two strategies: 
\begin{itemize}
    \item \textbf{Survey-based Evaluation:}
    A group of 7 learners participated in a structured, three-part process that included using LangLingual. The process was as follows:
    \begin{itemize}
        \item \textit{Pre-survey:} A web-based form is used to collect information on learners' English proficiency levels, learning goals and preferred learning styles.
        \item \textit{System Usage:} Learners interact with the LangLingual application, completing at least one full learning session.
        \item \textit{Post-survey:} A web-based form (separate from the previous) is used to collect feedback regarding learner experience, effectiveness of system features, and suggestions for future improvements.
    \end{itemize}
    \item \textbf{Persona-based Evaluation:} Three learner personas representing various English learning profiles were created. These personas were used to simulate common use-case scenarios, evaluating the system's contextual appropriateness, clarity of feedback and accuracy of responses. One of the authors of the paper assigned the scores based on pre-defined criteria. Persona-based testing of computer-based systems has been reported in the past~\cite{zanudin2021case, jandaghi2024faithful}
\end{itemize}

\subsection{Survey-based Evaluation}
\begin{figure}[htbp]
    \centering
    \includegraphics[width=1\textwidth]{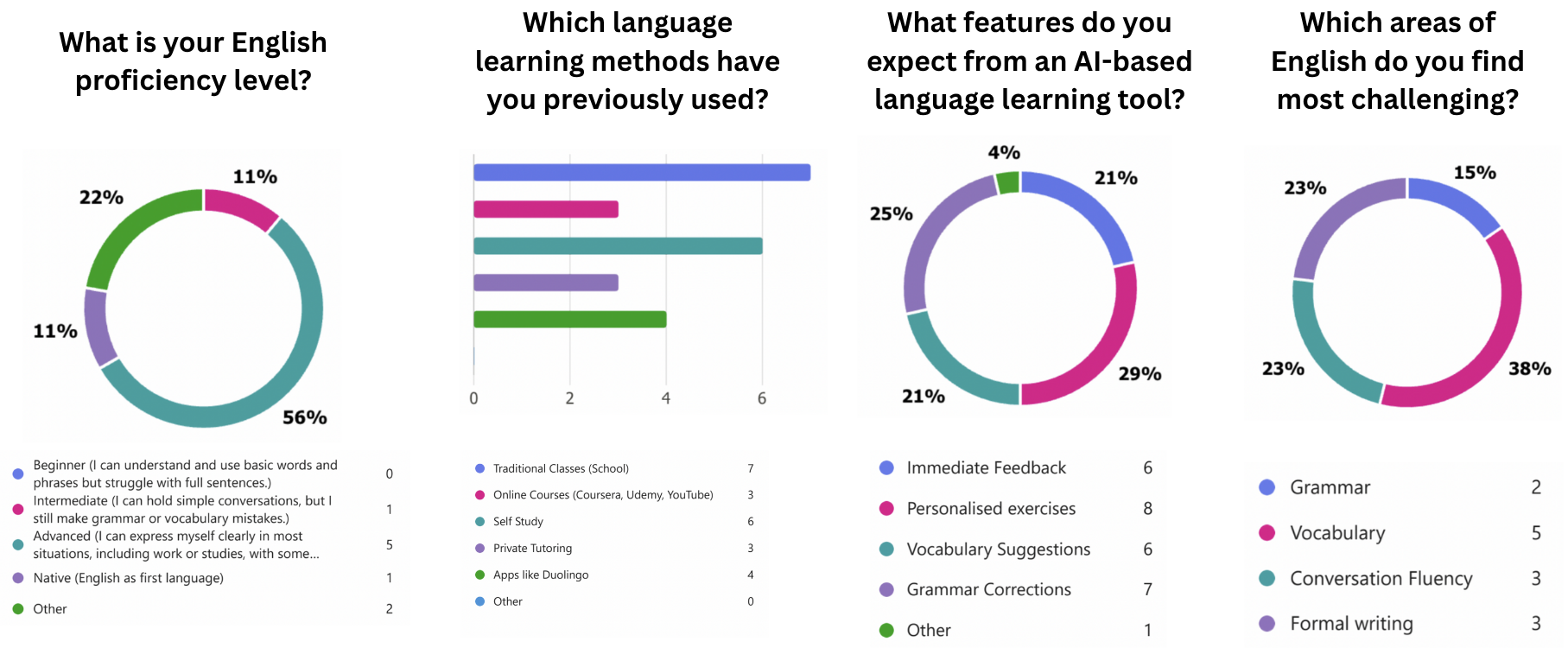}
    \caption{Findings from the pre-survey conducted to evaluate LangLingual.}
    \label{fig:pre-survey}
\end{figure}

We conduct two surveys to evaluate LangLingual. A \textbf{pre-survey} gathers insights into learners' English proficiency, motivation and challenges before using LangLingual. Following the pre-survey, learners interact with LangLingual, completing at least one full learning session. Subsequently, a \textbf{post-survey} collects feedback after interacting with LangLingual, focusing on perceived usefulness and efficacy of feedback. A total of \textbf{7 learners} participated in the surveys. All participants completed at least one full session with the app between the two surveys. The survey participants came from diverse backgrounds:

\begin{itemize}
    \item \textbf{Age Range} - 67\% of the participants were young adults in the 18-24 age range, with participants also in the ranges of 25-44.
    \item \textbf{Occupation} - The group included a mix of university students, working professionals and job seekers.
    \item \textbf{Nationality} - Participants came from a variety of national backgrounds, including India, Japan, Indonesia and Australia.
\end{itemize} 
Participation in the survey and the evaluation was completely voluntary. Participants were not authors of the paper.
\paragraph{Pre- survey}
The pre-survey has 12 questions in total with an average response time of 4.5 minutes. Figure~\ref{fig:pre-survey} shows responses for key questions in the pre-survey. Most of the participants rate themselves between intermediate and advanced proficiency levels, indicating a diverse but capable group of participants. Participants report using variety of language learning methods including online courses, private tutoring and apps such as DuoLingo. They report grammar correction and personalised exercise as two key features they would look for in a tool like LangLingual.
\FloatBarrier

Overall, the pre-survey insights confirmed that the test participants had identified their learning goals and were looking to improve specific skills in English. Their preferences supported the app's focus on personalised feedback, voice integration and improvement tracking.

\begin{figure}[htbp]
    \centering
    \includegraphics[width=0.9\textwidth]{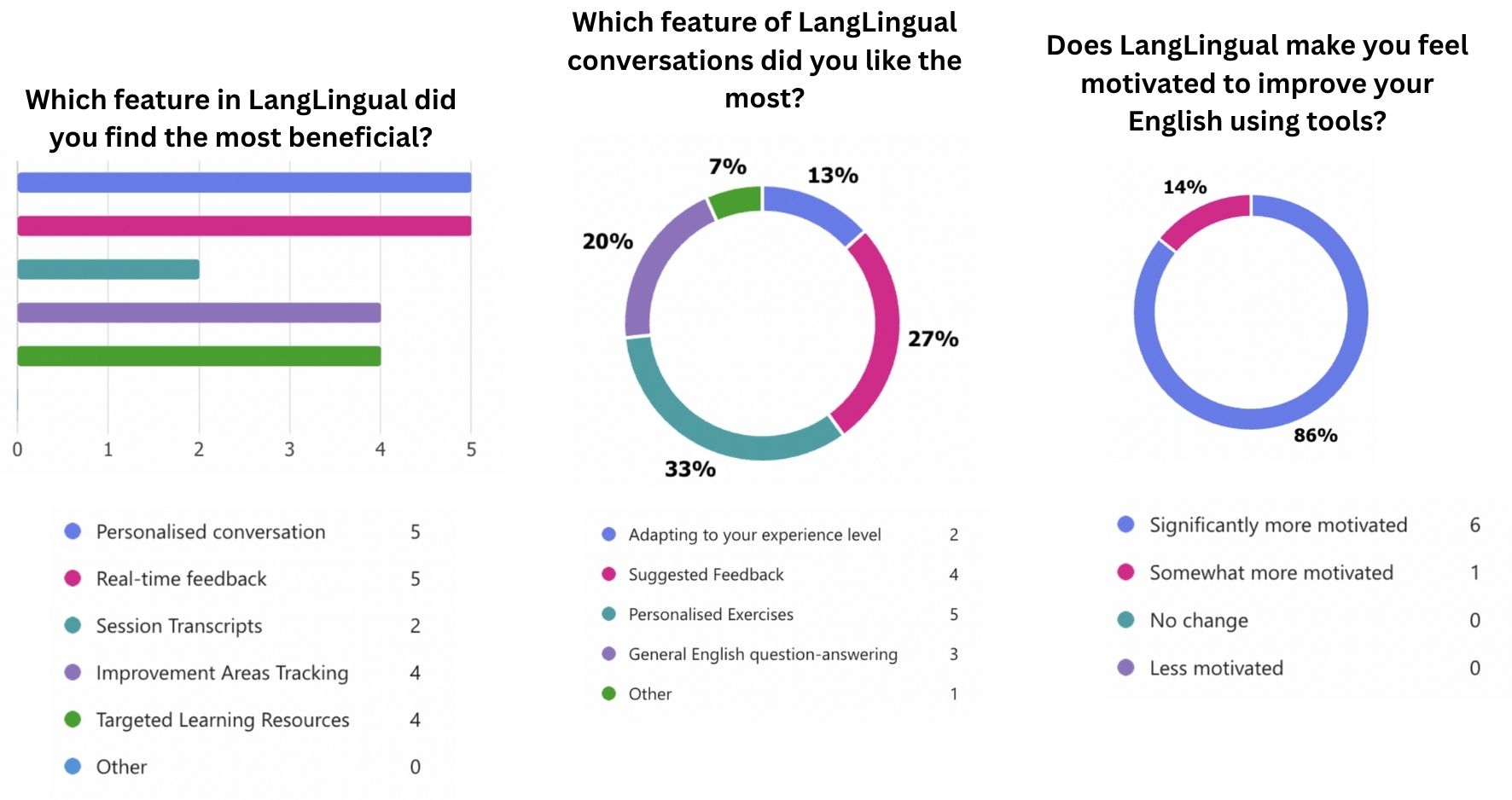}
    \caption{Findings from the ppost-survey conducted to evaluate LangLingual.}
    \label{fig:post-survey}
\end{figure}

\paragraph{Post- survey}
The post-survey had 14 questions with an average response time of 6.5 minutes. The post-survey was completed by participants after trying out LangLingual as described above. Post-survey responses, illustrated in Figure~\ref{fig:post-survey} show the quality of the learner experience and overall feedback for the application. Some visually representative responses can be seen below:

Figure~\ref{fig:post-survey} shows that most participants appreciate the personalised conversation and real-time\footnote{Real-time refers to immediate responses from the conversation agent.} feedback from LangLingual. This also resonated with the feature of LangLingual they like the most, being personalised exercises. Although the usage of LangLingual before completing the post-survey was not extensive, most participants report feeling significantly more motivated to use language learning tools.

Qualitative feedback from participants was also sought. The collected comments include statements such as:

\begin{itemize}
   \item  \emph{“I have noticed an improvement in the reduction of grammatical errors.”}
\item 
      \emph{“its ability to keep working with you on a particular aspect of your english until you get it right.”}
\end{itemize}

Overall, the post-survey results indicate that LangLingual provides an effective and personalised learning experience. Both learner-engagement and perceived skill improvement were positively rated, quantifying the success of the system.

\subsection{Persona-based Evaluation}
\begin{figure}[htbp]
    \centering
    \includegraphics[width=0.9\textwidth]{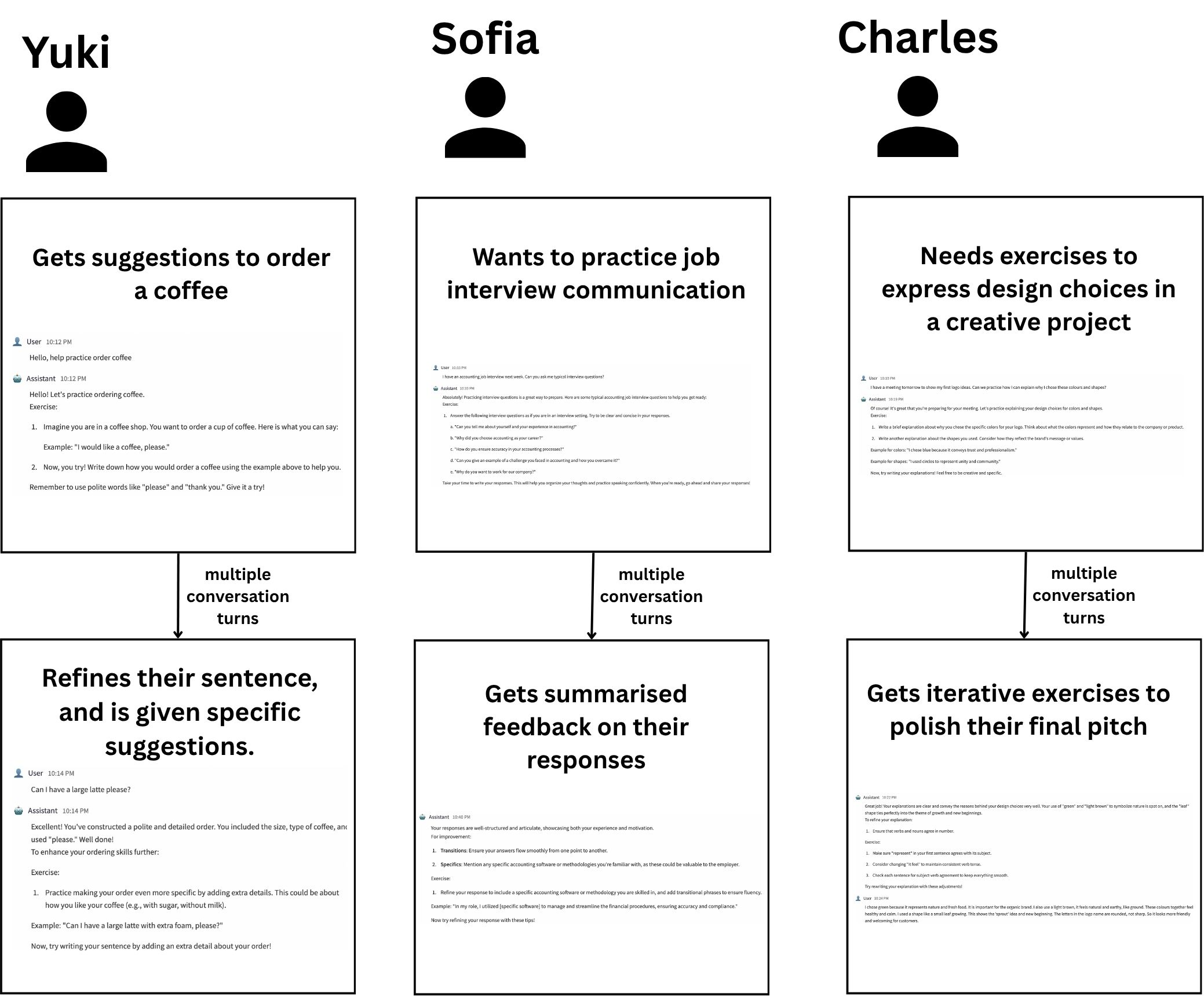}
   \caption{Persona-based evaluation}
    \label{fig:personas}
\end{figure}

To simulate diverse real-world usage, three learner personas were created based on varying English proficiency levels, context of usage and learner professions. This is similar to persona-based evaluation in general conversation systems ~\cite{jandaghi2024faithful, henka2014persona} and education-based conversation systems~\cite{stojmenova2013persona} such as ours. A human impersonates each of the personas based on their description and interacts with LangLingual. Figure~\ref{fig:personas} shows the interaction of the persona with LangLingual. The findings are qualitatively analysed as follows.

\paragraph{Persona 1: Yuki – Japanese University Student}
Yuki is a Japanese undergraduate student currently studying in Sydney. She is comfortable with basic English but lacks confidence in everyday spoken scenarios like ordering food or greeting strangers politely. Her goal is to practise simple, polite conversations to use in her daily life. She wants to practise everyday conversations like ordering coffee and greetings.  LangLingual guided her progression from an incomplete sentence to a polite and specific spoken request. It consistently provided short, focused exercises tailored to her learning level. It politely corrected grammar while maintaining conversational tone.

\paragraph{Persona 2: Sofia – Job-Seeking Accountant}
Sofia recently moved to Australia and is searching for an accounting job. She wants feedback on sounding fluent and convincing in interviews. She practises answering common interview questions and improving her confidence. LangLingual provided realistic interview questions and helped revise Sofia's responses. The feedback included grammar and fluency tips. The agent also provided tips to overcome nervousness with a template for practice.

\paragraph{Persona 3: Charles – Italian Graphic Designer}
Charles is a professional graphic designer preparing for a client meeting where he needs to present his first round of logo drafts. He has intermediate English skills but lacks confidence in expressing his ideas professionally  He wants to practice explaining design decisions and responding to spontaneous questions during meetings.

 LangLingual helped him with subject-verb agreement and verb tense consistency. It encouraged justification of his design decisions with examples.
Charles was able to practice professional responses using techniques like pausing, acknowledging  and following-up.

\section{Conclusion \& Future Work}\label{sec13}
LangLingual is a conversational agent to help English language learners by integrating Large Language Models (LLMs) with the LangChain framework\cite{langchain}. We designed, implemented and tested a working system called LangLingual, capable of generating personalised English Language exercises, detecting improvement areas and tracking learner proficiency. LangLingual successfully generates personalised English exercises based on learner context, history and proficiency level.
Furthermore, the agent identifies specific improvement areas for each learner session, helping learners prioritise their weakest aspects. It also enables learners to track their progress through a dashboard and allows them to view session summaries and download full transcripts for review.  learners receive actionable resources based on their detected improvement areas to support learning.
We evaluated using learner surveys and persona-based testing with positive results for engagement, usability and learning outcomes.

LangLingual can be improved in terms of proficiency Tracking. This may include pedagogy-informed thresholds for learners to advance to the next proficiency level, with rewards for maintaining consistent performance over multiple sessions.

LangLingual can be transformed from a reactive assistant to a proactive daily learning partner that encourages consistent engagement through regular check-ins and practice prompts. Similarly, learner retention can be improved by adding features such as streak tracking, daily notifications and gamified elements like badges and achievements.
We would also like to align proficiency levels in LangLingual with internationally recognised standards like the Common European Framework of Reference (CEFR) would also be instrumental.

\backmatter



\bigskip

\begin{appendices}




\end{appendices}

\bibliography{sn-bibliography}


\begin{thebibliography}{35}
\ifx \bisbn   \undefined \def \bisbn  #1{ISBN #1}\fi
\ifx \binits  \undefined \def \binits#1{#1}\fi
\ifx \bauthor  \undefined \def \bauthor#1{#1}\fi
\ifx \batitle  \undefined \def \batitle#1{#1}\fi
\ifx \bjtitle  \undefined \def \bjtitle#1{#1}\fi
\ifx \bvolume  \undefined \def \bvolume#1{\textbf{#1}}\fi
\ifx \byear  \undefined \def \byear#1{#1}\fi
\ifx \bissue  \undefined \def \bissue#1{#1}\fi
\ifx \bfpage  \undefined \def \bfpage#1{#1}\fi
\ifx \blpage  \undefined \def \blpage #1{#1}\fi
\ifx \burl  \undefined \def \burl#1{\textsf{#1}}\fi
\ifx \doiurl  \undefined \def \doiurl#1{\url{https://doi.org/#1}}\fi
\ifx \betal  \undefined \def \betal{\textit{et al.}}\fi
\ifx \binstitute  \undefined \def \binstitute#1{#1}\fi
\ifx \binstitutionaled  \undefined \def \binstitutionaled#1{#1}\fi
\ifx \bctitle  \undefined \def \bctitle#1{#1}\fi
\ifx \beditor  \undefined \def \beditor#1{#1}\fi
\ifx \bpublisher  \undefined \def \bpublisher#1{#1}\fi
\ifx \bbtitle  \undefined \def \bbtitle#1{#1}\fi
\ifx \bedition  \undefined \def \bedition#1{#1}\fi
\ifx \bseriesno  \undefined \def \bseriesno#1{#1}\fi
\ifx \blocation  \undefined \def \blocation#1{#1}\fi
\ifx \bsertitle  \undefined \def \bsertitle#1{#1}\fi
\ifx \bsnm \undefined \def \bsnm#1{#1}\fi
\ifx \bsuffix \undefined \def \bsuffix#1{#1}\fi
\ifx \bparticle \undefined \def \bparticle#1{#1}\fi
\ifx \barticle \undefined \def \barticle#1{#1}\fi
\bibcommenthead
\ifx \bconfdate \undefined \def \bconfdate #1{#1}\fi
\ifx \botherref \undefined \def \botherref #1{#1}\fi
\ifx \url \undefined \def \url#1{\textsf{#1}}\fi
\ifx \bchapter \undefined \def \bchapter#1{#1}\fi
\ifx \bbook \undefined \def \bbook#1{#1}\fi
\ifx \bcomment \undefined \def \bcomment#1{#1}\fi
\ifx \oauthor \undefined \def \oauthor#1{#1}\fi
\ifx \citeauthoryear \undefined \def \citeauthoryear#1{#1}\fi
\ifx \endbibitem  \undefined \def \endbibitem {}\fi
\ifx \bconflocation  \undefined \def \bconflocation#1{#1}\fi
\ifx \arxivurl  \undefined \def \arxivurl#1{\textsf{#1}}\fi
\csname PreBibitemsHook\endcsname

\bibitem[\protect\citeauthoryear{Mohamed et~al.}{2024}]{Mohamed2024-ax}
\begin{botherref}
\oauthor{\bsnm{Mohamed}, \binits{A.M.}},
\oauthor{\bsnm{Shaaban}, \binits{T.S.}},
\oauthor{\bsnm{Bakry}, \binits{S.H.}},
\oauthor{\bsnm{Guill{\'e}n-G{\'a}mez}, \binits{F.D.}},
\oauthor{\bsnm{Strzelecki}, \binits{A.}}:
Empowering the faculty of education students: Applying {AI's} potential for
  motivating and enhancing learning.
Innov. High. Educ.
(2024)
\end{botherref}
\endbibitem

\bibitem[\protect\citeauthoryear{Wambsganss
  et~al.}{2022}]{wambsganss-etal-2022-alen}
\begin{bchapter}
\bauthor{\bsnm{Wambsganss}, \binits{T.}},
\bauthor{\bsnm{Caines}, \binits{A.}},
\bauthor{\bsnm{Buttery}, \binits{P.}}:
\bctitle{{ALEN} app: Argumentative writing support to foster {E}nglish language
  learning}.
In: \bbtitle{Proceedings of the 17th Workshop on Innovative Use of NLP for
  Building Educational Applications (BEA 2022)},
pp. \bfpage{134}--\blpage{140}.
\bpublisher{Association for Computational Linguistics},
\blocation{Seattle, Washington}
(\byear{2022}).
\doiurl{10.18653/v1/2022.bea-1.18} .
\burl{https://aclanthology.org/2022.bea-1.18}
\end{bchapter}
\endbibitem

\bibitem[\protect\citeauthoryear{Mahajan}{2022}]{mahajan-2022-bela}
\begin{bchapter}
\bauthor{\bsnm{Mahajan}, \binits{M.}}:
\bctitle{{BELA}: Bot for {E}nglish language acquisition}.
In: \bbtitle{Proceedings of the Second Workshop on NLP for Positive Impact
  (NLP4PI)},
pp. \bfpage{142}--\blpage{148}.
\bpublisher{Association for Computational Linguistics},
\blocation{Abu Dhabi, United Arab Emirates (Hybrid)}
(\byear{2022}).
\doiurl{10.18653/v1/2022.nlp4pi-1.17} .
\burl{https://aclanthology.org/2022.nlp4pi-1.17}
\end{bchapter}
\endbibitem

\bibitem[\protect\citeauthoryear{Papariello}{2024}]{papluca/xlm-roberta}
\begin{botherref}
\oauthor{\bsnm{Papariello}, \binits{L.}}:
papluca/xlm-roberta-base-language-detection · Hugging Face
(2024).
\url{https://huggingface.co/papluca/xlm-roberta-base-language-detection}
\end{botherref}
\endbibitem

\bibitem[\protect\citeauthoryear{Madnani
  et~al.}{2016}]{madnani-etal-2016-language}
\begin{bchapter}
\bauthor{\bsnm{Madnani}, \binits{N.}},
\bauthor{\bsnm{Burstein}, \binits{J.}},
\bauthor{\bsnm{Sabatini}, \binits{J.}},
\bauthor{\bsnm{Biggers}, \binits{K.}},
\bauthor{\bsnm{Andreyev}, \binits{S.}}:
\bctitle{Language muse: Automated linguistic activity generation for {E}nglish
  language learners}.
In: \bbtitle{Proceedings of {ACL}-2016 System Demonstrations},
pp. \bfpage{79}--\blpage{84}.
\bpublisher{Association for Computational Linguistics},
\blocation{Berlin, Germany}
(\byear{2016}).
\doiurl{10.18653/v1/P16-4014} .
\burl{https://aclanthology.org/P16-4014}
\end{bchapter}
\endbibitem

\bibitem[\protect\citeauthoryear{Liang et~al.}{2023}]{liang-etal-2023-chatback}
\begin{bchapter}
\bauthor{\bsnm{Liang}, \binits{K.-H.}},
\bauthor{\bsnm{Davidson}, \binits{S.}},
\bauthor{\bsnm{Yuan}, \binits{X.}},
\bauthor{\bsnm{Panditharatne}, \binits{S.}},
\bauthor{\bsnm{Chen}, \binits{C.-Y.}},
\bauthor{\bsnm{Shea}, \binits{R.}},
\bauthor{\bsnm{Pham}, \binits{D.}},
\bauthor{\bsnm{Tan}, \binits{Y.}},
\bauthor{\bsnm{Voss}, \binits{E.}},
\bauthor{\bsnm{Fryer}, \binits{L.}}:
\bctitle{{C}hat{B}ack: Investigating methods of providing grammatical error
  feedback in a {GUI}-based language learning chatbot}.
In: \bbtitle{Proceedings of the 18th Workshop on Innovative Use of NLP for
  Building Educational Applications (BEA 2023)},
pp. \bfpage{83}--\blpage{99}.
\bpublisher{Association for Computational Linguistics},
\blocation{Toronto, Canada}
(\byear{2023}).
\doiurl{10.18653/v1/2023.bea-1.7} .
\burl{https://aclanthology.org/2023.bea-1.7}
\end{bchapter}
\endbibitem

\bibitem[\protect\citeauthoryear{Tyen et~al.}{2022}]{tyen-etal-2022-towards}
\begin{bchapter}
\bauthor{\bsnm{Tyen}, \binits{G.}},
\bauthor{\bsnm{Brenchley}, \binits{M.}},
\bauthor{\bsnm{Caines}, \binits{A.}},
\bauthor{\bsnm{Buttery}, \binits{P.}}:
\bctitle{Towards an open-domain chatbot for language practice}.
In: \bbtitle{Proceedings of the 17th Workshop on Innovative Use of NLP for
  Building Educational Applications (BEA 2022)},
pp. \bfpage{234}--\blpage{249}.
\bpublisher{Association for Computational Linguistics},
\blocation{Seattle, Washington}
(\byear{2022}).
\doiurl{10.18653/v1/2022.bea-1.28} .
\burl{https://aclanthology.org/2022.bea-1.28}
\end{bchapter}
\endbibitem

\bibitem[\protect\citeauthoryear{of~Europe}{2001}]{cefr}
\begin{botherref}
\oauthor{\bsnm{Europe}, \binits{C.}}:
Common European Framework of Reference for Languages: Learning, Teaching,
  Assessment (CEFR) - Common European Framework of Reference for Languages
  (CEFR) - www.coe.int
(2001).
\url{https://www.coe.int/en/web/common-european-framework-reference-languages}
\end{botherref}
\endbibitem

\bibitem[\protect\citeauthoryear{Roller
  et~al.}{2021}]{roller-etal-2021-recipes}
\begin{bchapter}
\bauthor{\bsnm{Roller}, \binits{S.}},
\bauthor{\bsnm{Dinan}, \binits{E.}},
\bauthor{\bsnm{Goyal}, \binits{N.}},
\bauthor{\bsnm{Ju}, \binits{D.}},
\bauthor{\bsnm{Williamson}, \binits{M.}},
\bauthor{\bsnm{Liu}, \binits{Y.}},
\bauthor{\bsnm{Xu}, \binits{J.}},
\bauthor{\bsnm{Ott}, \binits{M.}},
\bauthor{\bsnm{Smith}, \binits{E.M.}},
\bauthor{\bsnm{Boureau}, \binits{Y.-L.}},
\bauthor{\bsnm{Weston}, \binits{J.}}:
\bctitle{Recipes for building an open-domain chatbot}.
In: \bbtitle{Proceedings of the 16th Conference of the European Chapter of the
  Association for Computational Linguistics: Main Volume},
pp. \bfpage{300}--\blpage{325}.
\bpublisher{Association for Computational Linguistics},
\blocation{Online}
(\byear{2021}).
\doiurl{10.18653/v1/2021.eacl-main.24} .
\burl{https://aclanthology.org/2021.eacl-main.24}
\end{bchapter}
\endbibitem

\bibitem[\protect\citeauthoryear{Fan et~al.}{2018}]{fan-etal-2018-hierarchical}
\begin{bchapter}
\bauthor{\bsnm{Fan}, \binits{A.}},
\bauthor{\bsnm{Lewis}, \binits{M.}},
\bauthor{\bsnm{Dauphin}, \binits{Y.}}:
\bctitle{Hierarchical neural story generation}.
In: \bbtitle{Proceedings of the 56th Annual Meeting of the Association for
  Computational Linguistics (Volume 1: Long Papers)},
pp. \bfpage{889}--\blpage{898}.
\bpublisher{Association for Computational Linguistics},
\blocation{Melbourne, Australia}
(\byear{2018}).
\doiurl{10.18653/v1/P18-1082} .
\burl{https://aclanthology.org/P18-1082}
\end{bchapter}
\endbibitem

\bibitem[\protect\citeauthoryear{Sennrich
  et~al.}{2016}]{sennrich-etal-2016-neural}
\begin{bchapter}
\bauthor{\bsnm{Sennrich}, \binits{R.}},
\bauthor{\bsnm{Haddow}, \binits{B.}},
\bauthor{\bsnm{Birch}, \binits{A.}}:
\bctitle{Neural machine translation of rare words with subword units}.
In: \beditor{\bsnm{Erk}, \binits{K.}},
\beditor{\bsnm{Smith}, \binits{N.A.}} (eds.)
\bbtitle{Proceedings of the 54th Annual Meeting of the Association for
  Computational Linguistics (Volume 1: Long Papers)},
pp. \bfpage{1715}--\blpage{1725}.
\bpublisher{Association for Computational Linguistics},
\blocation{Berlin, Germany}
(\byear{2016}).
\doiurl{10.18653/v1/P16-1162} .
\burl{https://aclanthology.org/P16-1162}
\end{bchapter}
\endbibitem

\bibitem[\protect\citeauthoryear{Shatz}{}]{efcamdat-modified}
\begin{botherref}
\oauthor{\bsnm{Shatz}, \binits{I.}}:
Refining and modifying the EFCAMDAT: Lessons from creating a new corpus from an
  existing large-scale English learner language database.
\url{https://www.jbe-platform.com/content/journals/10.1075/ijlcr.20009.sha}
\end{botherref}
\endbibitem

\bibitem[\protect\citeauthoryear{Yannakoudakis et~al.}{}]{clc-fce}
\begin{botherref}
\oauthor{\bsnm{Yannakoudakis}, \binits{H.}},
\oauthor{\bsnm{Briscoe}, \binits{T.}},
\oauthor{\bsnm{Medlock}, \binits{B.}}:
A New Dataset and Method for Automatically Grading ESOL Texts.
\url{https://ilexir.co.uk/datasets/index.html}
\end{botherref}
\endbibitem

\bibitem[\protect\citeauthoryear{Tack et~al.}{2023}]{tack-etal-2023-bea}
\begin{bchapter}
\bauthor{\bsnm{Tack}, \binits{A.}},
\bauthor{\bsnm{Kochmar}, \binits{E.}},
\bauthor{\bsnm{Yuan}, \binits{Z.}},
\bauthor{\bsnm{Bibauw}, \binits{S.}},
\bauthor{\bsnm{Piech}, \binits{C.}}:
\bctitle{The {BEA} 2023 shared task on generating {AI} teacher responses in
  educational dialogues}.
In: \bbtitle{Proceedings of the 18th Workshop on Innovative Use of NLP for
  Building Educational Applications (BEA 2023)},
pp. \bfpage{785}--\blpage{795}.
\bpublisher{Association for Computational Linguistics},
\blocation{Toronto, Canada}
(\byear{2023}).
\doiurl{10.18653/v1/2023.bea-1.64} .
\burl{https://aclanthology.org/2023.bea-1.64}
\end{bchapter}
\endbibitem

\bibitem[\protect\citeauthoryear{Radford et~al.}{2019}]{radford2019language}
\begin{botherref}
\oauthor{\bsnm{Radford}, \binits{A.}},
\oauthor{\bsnm{Wu}, \binits{J.}},
\oauthor{\bsnm{Child}, \binits{R.}},
\oauthor{\bsnm{Luan}, \binits{D.}},
\oauthor{\bsnm{Amodei}, \binits{D.}},
\oauthor{\bsnm{Sutskever}, \binits{I.}}:
Language Models are Unsupervised Multitask Learners
(2019)
\end{botherref}
\endbibitem

\bibitem[\protect\citeauthoryear{Brown
  et~al.}{2020}]{brown2020languagemodelsfewshotlearners}
\begin{botherref}
\oauthor{\bsnm{Brown}, \binits{T.B.}},
\oauthor{\bsnm{Mann}, \binits{B.}},
\oauthor{\bsnm{Ryder}, \binits{N.}},
\oauthor{\bsnm{Subbiah}, \binits{M.}},
\oauthor{\bsnm{Kaplan}, \binits{J.}},
\oauthor{\bsnm{Dhariwal}, \binits{P.}},
\oauthor{\bsnm{Neelakantan}, \binits{A.}},
\oauthor{\bsnm{Shyam}, \binits{P.}},
\oauthor{\bsnm{Sastry}, \binits{G.}},
\oauthor{\bsnm{Askell}, \binits{A.}},
\oauthor{\bsnm{Agarwal}, \binits{S.}},
\oauthor{\bsnm{Herbert-Voss}, \binits{A.}},
\oauthor{\bsnm{Krueger}, \binits{G.}},
\oauthor{\bsnm{Henighan}, \binits{T.}},
\oauthor{\bsnm{Child}, \binits{R.}},
\oauthor{\bsnm{Ramesh}, \binits{A.}},
\oauthor{\bsnm{Ziegler}, \binits{D.M.}},
\oauthor{\bsnm{Wu}, \binits{J.}},
\oauthor{\bsnm{Winter}, \binits{C.}},
\oauthor{\bsnm{Hesse}, \binits{C.}},
\oauthor{\bsnm{Chen}, \binits{M.}},
\oauthor{\bsnm{Sigler}, \binits{E.}},
\oauthor{\bsnm{Litwin}, \binits{M.}},
\oauthor{\bsnm{Gray}, \binits{S.}},
\oauthor{\bsnm{Chess}, \binits{B.}},
\oauthor{\bsnm{Clark}, \binits{J.}},
\oauthor{\bsnm{Berner}, \binits{C.}},
\oauthor{\bsnm{McCandlish}, \binits{S.}},
\oauthor{\bsnm{Radford}, \binits{A.}},
\oauthor{\bsnm{Sutskever}, \binits{I.}},
\oauthor{\bsnm{Amodei}, \binits{D.}}:
Language Models are Few-Shot Learners
(2020).
\url{https://arxiv.org/abs/2005.14165}
\end{botherref}
\endbibitem

\bibitem[\protect\citeauthoryear{Reimers and
  Gurevych}{2019}]{reimers-gurevych-2019-sentence}
\begin{bchapter}
\bauthor{\bsnm{Reimers}, \binits{N.}},
\bauthor{\bsnm{Gurevych}, \binits{I.}}:
\bctitle{Sentence-{BERT}: Sentence embeddings using {S}iamese {BERT}-networks}.
In: \bbtitle{Proceedings of the 2019 Conference on Empirical Methods in Natural
  Language Processing and the 9th International Joint Conference on Natural
  Language Processing (EMNLP-IJCNLP)},
pp. \bfpage{3982}--\blpage{3992}.
\bpublisher{Association for Computational Linguistics},
\blocation{Hong Kong, China}
(\byear{2019}).
\doiurl{10.18653/v1/D19-1410} .
\burl{https://aclanthology.org/D19-1410}
\end{bchapter}
\endbibitem

\bibitem[\protect\citeauthoryear{Lewis et~al.}{2020}]{Lewis_2020_RAG}
\begin{bchapter}
\bauthor{\bsnm{Lewis}, \binits{P.}},
\bauthor{\bsnm{Perez}, \binits{E.}},
\bauthor{\bsnm{Piktus}, \binits{A.}},
\bauthor{\bsnm{Petroni}, \binits{F.}},
\bauthor{\bsnm{Karpukhin}, \binits{V.}},
\bauthor{\bsnm{Goyal}, \binits{N.}},
\bauthor{\bsnm{K\"{u}ttler}, \binits{H.}},
\bauthor{\bsnm{Lewis}, \binits{M.}},
\bauthor{\bsnm{Yih}, \binits{W.-t.}},
\bauthor{\bsnm{Rockt\"{a}schel}, \binits{T.}},
\bauthor{\bsnm{Riedel}, \binits{S.}},
\bauthor{\bsnm{Kiela}, \binits{D.}}:
\bctitle{Retrieval-augmented generation for knowledge-intensive nlp tasks}.
In: \bbtitle{Proceedings of the 34th International Conference on Neural
  Information Processing Systems}.
\bsertitle{NIPS '20}.
\bpublisher{Curran Associates Inc.},
\blocation{Red Hook, NY, USA}
(\byear{2020})
\end{bchapter}
\endbibitem

\bibitem[\protect\citeauthoryear{Xu
  et~al.}{2024}]{Xu_2024:LLM_augmented_exercise_retrieval}
\begin{bchapter}
\bauthor{\bsnm{Xu}, \binits{A.}},
\bauthor{\bsnm{Monroe}, \binits{W.}},
\bauthor{\bsnm{Bicknell}, \binits{K.}}:
\bctitle{Large language model augmented exercise retrieval for personalized
  language learning}.
In: \bbtitle{Proceedings of the 14th Learning Analytics and Knowledge
  Conference}.
\bsertitle{LAK '24},
pp. \bfpage{284}--\blpage{294}.
\bpublisher{Association for Computing Machinery},
\blocation{New York, NY, USA}
(\byear{2024}).
\doiurl{10.1145/3636555.3636883} .
\burl{https://doi.org/10.1145/3636555.3636883}
\end{bchapter}
\endbibitem

\bibitem[\protect\citeauthoryear{Kohnke et~al.}{2025}]{Kohnke_2025_exploring}
\begin{barticle}
\bauthor{\bsnm{Kohnke}, \binits{L.}},
\bauthor{\bsnm{Zou}, \binits{D.}},
\bauthor{\bsnm{Su}, \binits{F.}}:
\batitle{Exploring the potential of genai for personalised english teaching:
  Learners' experiences and perceptions}.
\bjtitle{Computers and Education: Artificial Intelligence}
\bvolume{8},
\bfpage{100371}
(\byear{2025})
\doiurl{10.1016/j.caeai.2025.100371}
\end{barticle}
\endbibitem

\bibitem[\protect\citeauthoryear{Cui and
  Sachan}{2023}]{cui-sachan-2023-adaptive}
\begin{bchapter}
\bauthor{\bsnm{Cui}, \binits{P.}},
\bauthor{\bsnm{Sachan}, \binits{M.}}:
\bctitle{Adaptive and personalized exercise generation for online language
  learning}.
In: \beditor{\bsnm{Rogers}, \binits{A.}},
\beditor{\bsnm{Boyd-Graber}, \binits{J.}},
\beditor{\bsnm{Okazaki}, \binits{N.}} (eds.)
\bbtitle{Proceedings of the 61st Annual Meeting of the Association for
  Computational Linguistics (Volume 1: Long Papers)},
pp. \bfpage{10184}--\blpage{10198}.
\bpublisher{Association for Computational Linguistics},
\blocation{Toronto, Canada}
(\byear{2023}).
\doiurl{10.18653/v1/2023.acl-long.567} .
\burl{https://aclanthology.org/2023.acl-long.567/}
\end{bchapter}
\endbibitem

\bibitem[\protect\citeauthoryear{Kim et~al.}{2024}]{Kim_2024_exploring_LLMs}
\begin{bchapter}
\bauthor{\bsnm{Kim}, \binits{M.}},
\bauthor{\bsnm{Nallbani}, \binits{A.L.}},
\bauthor{\bsnm{Stovall}, \binits{A.R.}}:
\bctitle{Exploring llm-based chatbot for language learning and cultivation of
  growth mindset}.
In: \bbtitle{Extended Abstracts of the CHI Conference on Human Factors in
  Computing Systems}.
\bsertitle{CHI EA '24}.
\bpublisher{Association for Computing Machinery},
\blocation{New York, NY, USA}
(\byear{2024}).
\doiurl{10.1145/3613905.3648628} .
\burl{https://doi.org/10.1145/3613905.3648628}
\end{bchapter}
\endbibitem

\bibitem[\protect\citeauthoryear{Yadegaridehkordi
  et~al.}{2025}]{Yadegaridehkordi2025-av}
\begin{botherref}
\oauthor{\bsnm{Yadegaridehkordi}, \binits{E.}},
\oauthor{\bsnm{Foroughi}, \binits{B.}},
\oauthor{\bsnm{Ghobakhloo}, \binits{M.}}:
Factors affecting academic staff's willingness to use {ChatGPT} for teaching
  and learning: A {PLS-SEM} and {ANN} approach.
Innov. High. Educ.
(2025)
\end{botherref}
\endbibitem

\bibitem[\protect\citeauthoryear{Streamlit.io}{}]{streamlit}
\begin{botherref}
\oauthor{\bsnm{Streamlit.io}}:
Streamlit's Website.
\url{https://streamlit.io/}
\end{botherref}
\endbibitem

\bibitem[\protect\citeauthoryear{supabase.com}{}]{supabase}
\begin{botherref}
\oauthor{\bsnm{supabase.com}}:
Supabase's Website.
\url{https://supabase.com/}
\end{botherref}
\endbibitem

\bibitem[\protect\citeauthoryear{postgresql.org}{}]{postgres}
\begin{botherref}
\oauthor{\bsnm{postgresql.org}}:
PostgreSQL's Website.
\url{https://www.postgresql.org/}
\end{botherref}
\endbibitem

\bibitem[\protect\citeauthoryear{Chase}{2022}]{langchain}
\begin{botherref}
\oauthor{\bsnm{Chase}, \binits{H.}}:
LangChain
(2022).
\url{https://python.langchain.com/docs/introduction/}
\end{botherref}
\endbibitem

\bibitem[\protect\citeauthoryear{openai.com}{}]{openai}
\begin{botherref}
\oauthor{\bsnm{openai.com}}:
OpenAI's Website.
\url{https://openai.com/}
\end{botherref}
\endbibitem

\bibitem[\protect\citeauthoryear{openai.com}{}]{gpt-models}
\begin{botherref}
\oauthor{\bsnm{openai.com}}:
OpenAI's Models Documentation.
\url{https://platform.openai.com/docs/models}
\end{botherref}
\endbibitem

\bibitem[\protect\citeauthoryear{openai.com}{}]{whisper}
\begin{botherref}
\oauthor{\bsnm{openai.com}}:
OpenAI's Whisper Documentation.
\url{https://platform.openai.com/docs/models/whisper-1}
\end{botherref}
\endbibitem

\bibitem[\protect\citeauthoryear{trychroma.com}{}]{chroma}
\begin{botherref}
\oauthor{\bsnm{trychroma.com}}:
Chroma's Website.
\url{https://www.trychroma.com/}
\end{botherref}
\endbibitem

\bibitem[\protect\citeauthoryear{Zanudin et~al.}{2021}]{zanudin2021case}
\begin{bchapter}
\bauthor{\bsnm{Zanudin}, \binits{N.N.}},
\bauthor{\bsnm{Sulaiman}, \binits{S.}},
\bauthor{\bsnm{Samingan}, \binits{M.R.}},
\bauthor{\bsnm{Mohamed}, \binits{H.}},
\bauthor{\bsnm{Raof}, \binits{S.K.S.A.}},
\bauthor{\bsnm{Abd~Samad}, \binits{A.R.}}:
\bctitle{Case study on prototyping educational applications using persona-based
  approach}.
In: \bbtitle{2021 8th International Conference on Computer and Communication
  Engineering (ICCCE)},
pp. \bfpage{93}--\blpage{98}
(\byear{2021}).
\bcomment{IEEE}
\end{bchapter}
\endbibitem

\bibitem[\protect\citeauthoryear{Jandaghi et~al.}{2024}]{jandaghi2024faithful}
\begin{bchapter}
\bauthor{\bsnm{Jandaghi}, \binits{P.}},
\bauthor{\bsnm{Sheng}, \binits{X.}},
\bauthor{\bsnm{Bai}, \binits{X.}},
\bauthor{\bsnm{Pujara}, \binits{J.}},
\bauthor{\bsnm{Sidahmed}, \binits{H.}}:
\bctitle{Faithful persona-based conversational dataset generation with large
  language models}.
In: \bbtitle{Findings of the Association for Computational Linguistics ACL
  2024},
pp. \bfpage{15245}--\blpage{15270}
(\byear{2024})
\end{bchapter}
\endbibitem

\bibitem[\protect\citeauthoryear{Henka and Zimmermann}{2014}]{henka2014persona}
\begin{bchapter}
\bauthor{\bsnm{Henka}, \binits{A.}},
\bauthor{\bsnm{Zimmermann}, \binits{G.}}:
\bctitle{Persona based accessibility testing: Towards user-centered
  accessibility evaluation}.
In: \bbtitle{International Conference on Human-Computer Interaction},
pp. \bfpage{226}--\blpage{231}
(\byear{2014}).
\bcomment{Springer}
\end{bchapter}
\endbibitem

\bibitem[\protect\citeauthoryear{Stojmenova
  et~al.}{2013}]{stojmenova2013persona}
\begin{bchapter}
\bauthor{\bsnm{Stojmenova}, \binits{E.}},
\bauthor{\bsnm{Lugmayr}, \binits{A.}},
\bauthor{\bsnm{Dinevski}, \binits{D.}}:
\bctitle{Persona-based expert review of an e-learning system for adults}.
In: \bbtitle{2013 IEEE International Conference on Multimedia and Expo
  Workshops (ICMEW)},
pp. \bfpage{1}--\blpage{4}
(\byear{2013}).
\bcomment{IEEE}
\end{bchapter}
\endbibitem

\end{thebibliography}

\end{document}